\documentclass[acmtog]{acmart}
\acmSubmissionID{967}

\usepackage{booktabs} 

\usepackage{mmstyle}

\usepackage[ruled]{algorithm2e} 

\SetAlFnt{\small}
\SetAlCapFnt{\small}
\SetAlCapNameFnt{\small}
\SetAlCapHSkip{0pt}

\copyrightyear{2024}
\acmYear{2024}\setcopyright{acmlicensed}\acmConference[SA Conference Papers '24]{SIGGRAPH Asia
2024 Conference Papers}{December 3--6, 2024}{Tokyo, Japan}
\acmBooktitle{SIGGRAPH Asia 2024 Conference Papers (SA Conference Papers '24),
December 3--6, 2024, Tokyo, Japan}
\acmDOI{10.1145/3680528.3687676}
\acmISBN{979-8-4007-1131-2/24/12}

\begin{document}
\title{Boosting 3D Object Generation through PBR Materials}

\author{Yitong Wang}
\affiliation{%
  \institution{Fudan University}
  \city{Shanghai}
  \country{China}
}
\affiliation{%
  \institution{Shanghai Artificial Intelligence Laboratory}
  \city{Shanghai}
  \country{China}
}
\email{wangyitong23@m.fudan.edu.cn}

\author{Xudong Xu}
\affiliation{%
  \institution{Shanghai Artificial Intelligence Laboratory}
  \city{Shanghai}
  \country{China}
}
\email{xuxudong@pjlab.org.cn}

\author{Li Ma}
\affiliation{%
  \institution{Netflix Eyeline Studios}
  \city{Los Angeles}
  \country{United States of America}
}
\email{lmaag@connect.ust.hk}

\author{Haoran Wang}
\affiliation{%
  \institution{Shanghai Jiao Tong University}
  \city{Shanghai}
  \country{China}
}
\email{a.museum@sjtu.edu.cn}

\author{Bo Dai}
\affiliation{%
  \institution{Shanghai Artificial Intelligence Laboratory}
  \city{Shanghai}
  \country{China}
}
\email{daibo@pjlab.org.cn}

\begin{abstract}
Automatic 3D content creation has gained increasing attention recently, due to its potential in various applications such as video games, film industry, and AR/VR. 
Recent advancements in diffusion models and multimodal models have notably improved the quality and efficiency of 3D object generation given a single RGB image. 
However, 3D objects generated even by state-of-the-art methods are still unsatisfactory compared to human-created assets. 
Considering only textures instead of materials makes these methods encounter challenges in photo-realistic rendering, relighting, and flexible appearance editing.
And they also suffer from severe misalignment between geometry and high-frequency texture details. 
In this work, we propose a novel approach to boost the quality of generated 3D objects from the perspective of Physics-Based Rendering (PBR) materials. 
By analyzing the components of PBR materials, 
we choose to consider albedo, roughness, metalness, and bump \textcolor{black}{maps}. 
For albedo and bump maps,
we leverage Stable Diffusion fine-tuned on synthetic data to extract these values,
with novel usages of these fine-tuned models to obtain 3D consistent albedo UV and bump UV for generated objects.
In terms of roughness and metalness \textcolor{black}{maps},
we adopt a semi-automatic process to provide room for interactive adjustment, which we believe is more practical.
Extensive experiments demonstrate that our model is generally beneficial for various state-of-the-art generation methods, significantly boosting the quality and realism of their generated 3D objects, with natural relighting effects and substantially improved geometry.
\end{abstract}

\ccsdesc[300]{Computing methodologies~Appearance and texture representations}

%
%

\keywords{material generation, 3D asset creation, generative modeling}

\begin{teaserfigure}
  \vspace{-0.15in}
  \centering
  \includegraphics[width=17cm]{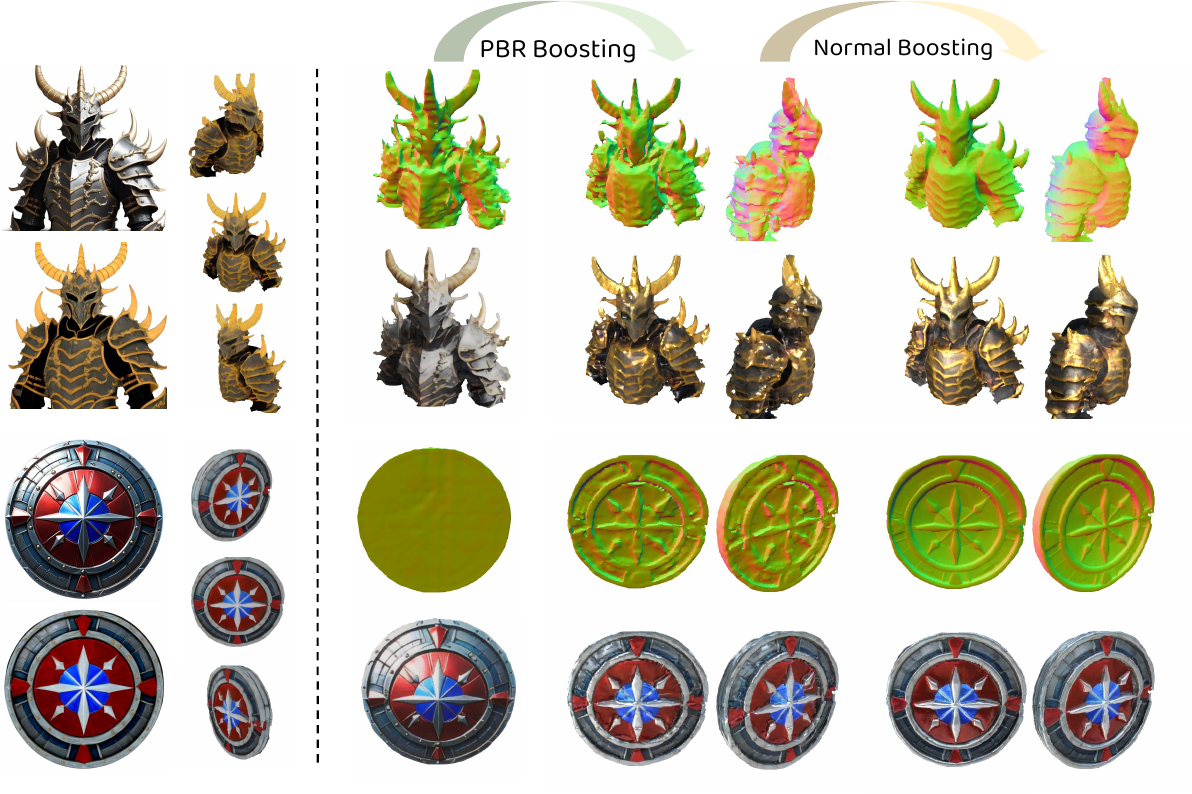}
  \vspace{-20pt}
  \caption{\small \textbf{Overview.} Given a single image, the existing image-to-3D generative models always synthesize 3D meshes with flawed geometry and RGB textures only.
  Our method not only boosts existing approaches with PBR materials, empowering relighting under various lighting conditions, but also boosts the object's normal maps, capturing more intricate details and better aligning with the given image.
  Notably, we fine-tune Stable Diffusion to estimate the albedo \textcolor{black}{map} from the single-view RGB image and lift it to multi-view albedo \textcolor{black}{maps} for a complete albedo UV.}
  \label{fig:teaser}
\end{teaserfigure}

\maketitle


\section{Introduction}

3D content generation has gathered widespread interest in recent years for its vast potential in diverse applications,
such as video games, filmmaking, and AR/VR.
The advancement of diffusion models~\cite{ho2020denoising,rombach2022high} has precipitated a paradigm shift in 3D content generation,
significantly enhancing the realism of produced 3D objects.
Owing to its unique fast feedforward pipeline and controllability,
3D from a single image gradually becomes the main pipeline in 3D content generation.
Given an RGB image and a conditional text prompt,
these methods~\cite{long2023wonder3d,wang2024crm,liu2023syncdreamer,xu2024instantmesh,hong2023lrm} can project the input image to multi-view images and then fuse them into a 3D mesh with compelling textures and great 3D consistency.

Notwithstanding the remarkable progress, existing methods still suffer from two fundamental drawbacks.
For one, all of them can only generate 3D objects with textures but ignore more crucial materials, which are imperative for rendering under various lighting conditions.
The absence of materials not only compromises the photorealism of 3D objects but also constrains their utility in a wider range of downstream applications.
For another, the generated 3D objects often exhibit a misalignment between their geometry and high-frequency details of corresponding textures, resulting in a modest geometry quality that falls short of expectations.
Even when endowed with plausible materials, these 3D objects tend to exhibit unrealistic artifacts under novel illuminations.

In this paper, we propose a novel approach to boost 3D object generation in the perspective of Physics-Based Rendering (PBR) materials. It works in a plug-and-play manner that is compatible with any single image-to-3D generation method.
By considering PBR materials,
objects generated with our approach can be more photorealistic and relightable thanks to the involvement of concepts like albedo, roughness, and metalness. 
Besides,
the misalignment of high-frequency geometry details can also be substantially improved since PBR materials cover bump maps that reflect intricate texture-aligned details.

In our proposed approach,
different components of PBR materials are handled in different ways to enhance their practical value,
where albedo maps are predicted from the input RGB image, bump maps are iteratively optimized given a 3D mesh and its albedo UV, and roughness and metalness \textcolor{black}{maps} are determined in a semi-automatic way, to leave space for interactive adjustment as desired by practical workflows.
Specifically,
to predict albedo maps and optimize bump maps, we fine-tune Stable Diffusion with synthetic data to obtain image-to-albedo and image-to-normal diffusion models,
motivated by the promising prospects of unleashing the diffusion priors for intrinsic properties.
Subsequently,
to obtain the albedo UV and bump UV of a target 3D object,
we first convert the input image into an albedo \textcolor{black}{map},
which is fed into an image-to-3D generation method to generate multi-view albedo \textcolor{black}{maps}.
we empirically found this leads to satisfactory results since albedo \textcolor{black}{maps} can be seen as clean images without much noise.
A 3D mesh with a complete albedo UV is then obtained by fusing these multi-view albedo \textcolor{black}{maps},
whose geometry contains severe misalignment as discussed above.
We thus further apply an iterative refinement process based on the previously fine-tuned image-to-normal diffusion model,
where we 
refine the original normals by optimizing bump UV from different viewing angles, until satisfactory normals are obtained.
While we can adopt a similar prediction process for roughness and metalness UV,
we argue that a semi-automatic process with interactive adjustment functionality is more preferred in real applications.
In our proposed approach,
such a semi-automatic process is achieved by 
leveraging the Segment-Anything-Model~\cite{kirillov2023segany} to obtain 3D segmentation masks indicating object regions that should be consistent in terms of semantics, as well as roughness and metalness values.
Afterwards,
the roughness and metalness values of each part can be recommended by powerful Vision-Language Models (VLMs)~\cite{achiam2023gpt,team2023gemini,liu2023llava} or manually adjusted by experienced 3D artists.

As shown in Figure~\ref{fig:teaser},
our fine-tuned diffusion models are capable of estimating accurate albedo and normal maps from a single RGB image.
Thanks to such superior performance,
our plug-and-play approach seamlessly integrates CRM~\cite{wang2024crm}, a state-of-the-art single image-to-3D generation approach,
to produce high-quality 3D objects with authentic PBR materials that support relighting under diverse illuminations.
Exhaustive experiments, as shown in Figure~\ref{fig:geo_boost4}, ~\ref{fig:edit} and~\ref{fig:pbr_compare}, demonstrate that
our model can substantially boost various 3D generation frameworks, efficiently yielding 3D assets with relightable capability and intricate normal details.

\section{Related Work}

\begin{figure*}[t!]
    \centering
    \includegraphics[width=0.98\linewidth]{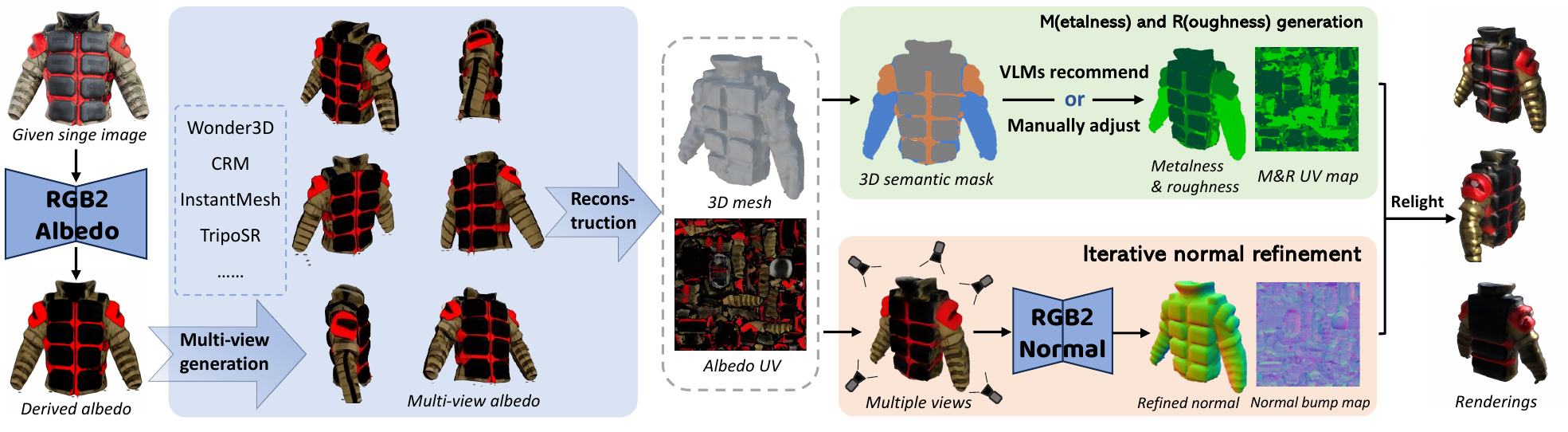}
    \vspace{-12pt}
    \caption{\small \textbf{Overview of our 3D generation pipeline.} Given a single image, we first convert it to an albedo \textcolor{black}{map} using our fine-tuned diffusion model. Conditioned on this derived albedo, the base method to be boosted will generate multi-view albedo \textcolor{black}{maps} and then fuse them into a 3D mesh and an albedo UV. Afterwards, we leverage a 3D semantic mask to obtain complete metalness and roughness UVs by acquiring the VLMs or 3D artists' manual adjustment. Moreover, an iterative normal refinement is employed to boost the original flawed normals, empowering realistic relighting results.}
    \label{fig:framework}
\end{figure*}

\paragraph{Text-to-3D generation with 2D diffusion models}
With the advancement of text-to-image diffusion models, a line of research work seeks to exploit strong priors from 2D diffusion models for 3D content generation.
Pioneers, 
DreamFusion~\cite{poole2022dreamfusion} and SJC~\cite{wang2023score}, propose Score Distillation Sampling (SDS) (also known as Score Jacobian Chaining) that significantly facilitates the development of this area.
Following the SDS-based 2D-lifting method, recent works have achieved promising results through improved score distillation loss~\cite{wang2023prolificdreamer}, texture refinement~\cite{chen2024it3d,liang2023luciddreamer}, multi-view diffusion model~\cite{shi2023mvdream}, or more advance 3D representations~\cite{li2023sweetdreamer}.
However, these methods solely generate 3D objects with RGB textures, devoid of materials, thereby failing to satisfy the requirements of real-world applications.
Fantasia3D~\cite{chen2023fantasia3d} tried to model PBR materials by disentangling geometry and texture but lacked the necessary constraints on PBR materials and illuminations to achieve such disentanglement.
RichDreamer~\cite{qiu2023richdreamer} and UniDream~\cite{liu2023unidream} both employ diffusion models trained on the albedo domain for PBR material modeling, but unfortunately, they inadvertently bake geometry details into the metalness and roughness components.
Unlike them, our approach leverages the diffusion priors for albedo estimation and incorporates 3D semantic masks for more plausible metalness and roughness maps.

\paragraph{Single image-to-3D generation}
By training on medium-sized 3D object datasets,
prior works have investigated image-conditioned 3D generative models on various 3D representations~\cite{melas2023pc2,liu2023meshdiffusion,cheng2023sdfusion,muller2023diffrf},
but the diversity and generalization of their produced 3D objects are significantly limited.
The seminal work, Zero123~\cite{liu2023zero}, fine-tunes a view-conditioned diffusion model on a large-scale 3D dataset, Objaverse~\cite{deitke2023objaverse}, for novel view synthesis, inspiring plenty of optimization-based approaches~\cite{qian2023magic123,lin2023consistent123,sun2023dreamcraft3d}.
Meanwhile,
some works~\cite{liu2023one,liu2023syncdreamer,long2023wonder3d} design a more promising reconstruction-based paradigm that consists of multi-view image generation and 3D reconstruction from these views.
Given a single image,
Wonder3D~\cite{long2023wonder3d} generates 6 novel views and the corresponding normal maps, which are then fused to a textured 3D mesh via NeuS~\cite{wang2021neus}.
Recently, LRM-series methods~\cite{hong2023lrm,li2023instant3d,wang2023pf,xu2023dmv3d,tochilkin2024triposr,wang2024crm,xu2024instantmesh} have received increasing attention due to their distinctive feed-forward architecture and rapid reconstruction speeds.
However, all these image-to-3D methods ignore PBR materials and their produced 3D objects lack geometry details.
we aim to boost all of them with realistic PBR materials and refine their created 3D assets with intricate normals.

\paragraph{Material Capture and Generation}
Thanks to the advent of deep learning,
single flash image material estimation~\cite{hui2017reflectance,kang2018efficient,deschaintre2018single,guo2020materialgan,gao2019deepbrdf} has already made great progress by leveraging U-Net architecture~\cite{ronneberger2015u}, albeit under the assumption of 2D planar geometry.
Moreover,
prior endeavors~\cite{forsyth2021intrinsic,wimbauer2022rendering,sang2020single} aimed at recovering materials from the in-the-wild single RGB image rely on feed-forward neural networks,
whereas we target unleashing 2D diffusion priors to estimate albedo instead.
On the other hand,
some methods\textcolor{black}{~\cite{vecchio2023controlmat,lopes2023material,matfusion,RGBtoX,kocsis2024iid}} utilize diffusion-based generative models to synthesize material conditioned on input photographs.
Similar to ours,
Another body of research works~\cite{xu2023matlaber,youwang2023paint,chen2022tango,vainer2024collaborative} concentrates on generating 3D meshes with PBR materials, but they either require extensive training or cannot achieve satisfactory material disentanglement from illuminations.
In contrast,
our method enhances overall fidelity and realism with realistic PBR materials where the normal maps help to recover more geometry details.


\section{Preliminaries}

\subsection{Stable Diffusion}
Stable Diffusion is a latent diffusion model \cite{rombach2022stablediffusion} which has achieved state-of-the-art performance in text-to-image generation.
It performs the diffusion process in latent space to enable the generation of high-resolution images. A variational autoencoder (VAE) is used to decode and encode the image to and from the latent space.
The crux of the diffusion process is a U-Net that predicts the noise $\hat{\epsilon}$ from a noisy latent $\mathbf{z}_t$, given a text embedding $\mathbf{s}$ and the timestep $t$: $\hat{\epsilon} = g(\mathbf{z}_t; \mathbf{s}, t),$
where $g$ represents the function modeled by the U-Net.
By iteratively removing the noise from an initial random noise, a clean latent $\mathbf{z}_0$ is generated, which can then be decoded into the resulting image. 

Existing works have demonstrated that pre-trained Stable Diffusion can serve as a vision foundation model. After fine-tuning, they can be adapted for various down-streaming vision tasks, such as relighting \cite{kocsis2024lightit}, human reenactment \cite{hu2023animateanyone}, image editing \cite{huang2024diffusion}, and depth estimation \cite{ke2023repurposing}. In this work, we exploit pre-trained stable diffusion as a prior model for predicting albedo and normal maps from a single image.

\subsection{3D Reconstruction from a Single Image}
The reconstruction-based methods for recovering 3D meshes from single images typically involve a two-stage pipeline, comprising the generation of multi-view images and the subsequent reconstruction of 3D geometry from these synthesized views.
Given a single image $I_0\in \mathbb{R}^{H\times W\times C}$,
these approaches employ a multi-view diffusion model $\cG_M$ to generate a set of consistent multi-view images:
\begin{equation}
    \label{eq:gen_mv}
    I_{1:N} = \cG_M(I_0).
\end{equation}

Thereafter, these multi-view images $I_{1:N}$ are fused into a 3D mesh $M$ with accompanying textures $T$.
These methods either leverage sparse-view 3D reconstruction algorithms such as NeuS~\cite{wang2021neus} or train a fast feed-forward reconstruction model~\cite{hong2023lrm}.
While sparse-view reconstruction algorithms require a significant amount of time to process each object, feed-forward methods, which are trained on large-scale 3D datasets, have demonstrated exceptional speed and generalization capabilities.
Denote the reconstruction model as $\cG_R$, we formulate this process as
\begin{equation}
    \label{eq:fuse}
    (M, T) = \cG_R(I_{1:N}),
\end{equation}
where the texture $T$ can be further expanded to a UV map.

\section{Methodology}
This section elaborates on our plug-and-play method that boosts single image-to-3D generation frameworks through PBR materials.
The overview of our whole pipeline is illustrated in Figure~\ref{fig:framework}.
Section~\ref{sec:an_est} first provides details on the fine-tuning of Stable Diffusion to accurately estimate the albedo and normal map from a given RGB image.
Following the estimation of the albedo, we elaborate in Section~\ref{sec:pbr} on how to leverage the Vision-Language Models (VLMs) to assign plausible values for metalness and roughness terms, with the guidance of 3D semantic masks.
Finally, we propose the iterative normal refinement in Section~\ref{sec:refine} where the derived normal maps in Section~\ref{sec:an_est} are treated as the pseudo-ground truth.

\subsection{Albedo and Normal Estimation}
\label{sec:an_est}
In our pipeline, two image-to-image translation modules are employed to predict the albedo and normal map, respectively, from a single input image. 
However, estimating albedo or normal maps from a single image is a highly ill-posed problem due to the lack of lighting or geometry information. Therefore, a strong prior is essential to recover plausible albedo and normal maps from a single image. Inspired by existing work on monocular depth estimation \cite{ke2023repurposing}, we exploit the data-driving prior inside the Stable Diffusion \cite{rombach2022stablediffusion} to achieve zero-shot albedo and normal map estimation. We show that by slightly modifying the U-Net structure and fine-tuning the pre-trained stable diffusion model on the synthetic dataset, we can obtain an image-to-image translation model that generalizes well to unseen in-the-wild data.
It is noteworthy that such an image-to-image translation paradigm leads to high-quality albedo \textcolor{black}{maps} without clear highlights or shadows and intricate normals with fine details.

To this end, we initially encode the input single image using the VAE encoder $\cE$ into a latent code $\mathbf{z}_i$.
Then, we concatenate the input latent with the noisy latent and feed the resulting composite latent code into the U-Net of Stable Diffusion:
\begin{equation}
    \hat{\epsilon}_{\text{task}} = g_{\text{task}}(\mathbf{z}_t \parallel \mathbf{z}_i; \mathbf{s}_\emptyset, t), \text{task} \in \{\text{normal}, \text{albedo}\},
    \label{eq:our_sd}
\end{equation}
where $\parallel$ is the concatenation operator, and $\mathbf{s}_\emptyset$ indicates an empty text embedding. 
Note that the U-Net $g$ is originally designed to take in the noisy latent only. Therefore, we duplicate the number of input channels for the first convolutional layer inside U-Net to enable the concatenated latent. The U-Net gradually denoises the noisy latent into a clean latent, which is then decoded into a normal map or an albedo map using the VAE decoder.
Theoretically, we can control whether the U-Net generates the normal or albedo map based on the text prompt $\mathbf{s}$. However, we observe that sharing the same network for different tasks leads to slightly degenerate predictions. Therefore, we train two separate U-Net, $g_\text{normal}$ and $g_\text{albedo}$, for albedo and normal estimation, and leave the text embedding as an empty text embedding $\mathbf{s}_\emptyset$.


\subsection{PBR Material Generation}
\label{sec:pbr}
To obtain a complete albedo UV,
a straightforward solution is to utilize our fine-tuned image-to-albedo diffusion model to derive multi-view albedo \textcolor{black}{maps} from generated multi-view images $I_{1:N}$ as described in Equation~(\ref{eq:gen_mv}).
However, we empirically find such a naive approach results in inconsistent albedo \textcolor{black}{maps}.
Instead, we first leverage the diffusion model to convert the given single image to the albedo \textcolor{black}{map} following Equation~(\ref{eq:our_sd}),
and then employ Equation~(\ref{eq:gen_mv}) to synthesize multi-view albedo \textcolor{black}{maps} conditioned on this derived albedo.
The multi-view albedo \textcolor{black}{map} can be fused to a 3D mesh $M$ and an albedo UV $A$ via Equation~(\ref{eq:fuse}).

While generating the metalness and roughness maps,
we conform to the inherent property of PBR materials, \ie, surface areas with similar semantic characteristics tend to exhibit consistent values.
Specifically,
we project the reconstructed 3D mesh from 6 orthographic views and obtain 6 orthographic albedo \textcolor{black}{maps}, which are segmented into different parts via the Segment-Anything-Model~\cite{kirillov2023segany}.
Through voting strategy in the overlapping regions,
such six segmentation results can be seamlessly integrated into a 3D semantic mask, as illustrated in Figure~\ref{fig:framework}.
Thereafter, we feed the given image into Gemini~\cite{team2023gemini}, one of the powerful Vision-Language models, to get the recommended values of metalness and roughness terms associated with different semantic parts.
Equipped with this 3D mask, we can easily extend these values to the entire 3D object, thereby generating comprehensive metalness and roughness UVs.
Moreover, it's noteworthy that the values of such two terms are typically adjusted by experienced 3D artists in practical 3D content creation workflows.

\subsection{Iterative Normal Refinement}
\label{sec:refine}
Unfortunately, the normal map of reconstructed 3D meshes contains too many flaws, leading to poor relighting results as shown in Figure~\ref{fig:geo_relight}.
To overcome this challenge, we propose iterative normal refinement by using the aforementioned normal estimation diffusion model.

We draw inspiration from the texture refinement presented in DreamGaussian~\cite{tang2023dreamgaussian} and propose a refinement strategy involving optimizing a bump map,
which combines with the original flawed normal $n_o$ to produce a refined normal map $n_f(\theta)$.
Specifically, an MLP $\Gamma$ parameterized as $\theta$ is utilized to predict the bump map $n_b(\theta)$.
For any point $p\in R^3$ on the surface of a 3D mesh $M$, we apply the hash-grid positional encoding $\beta(\cdot)$ on point $p$ and then obtain the bump map and refined normal map via:
\begin{equation}
    n_b(\theta) = \Gamma(\beta(p) ; \theta), \quad  n_f(\theta) = n_o \oplus n_b(\theta),
\end{equation}
where $\oplus$ represents the special operation for normal integration.
\begin{figure*}[t!]
    \centering
    \includegraphics[width=0.98\linewidth]{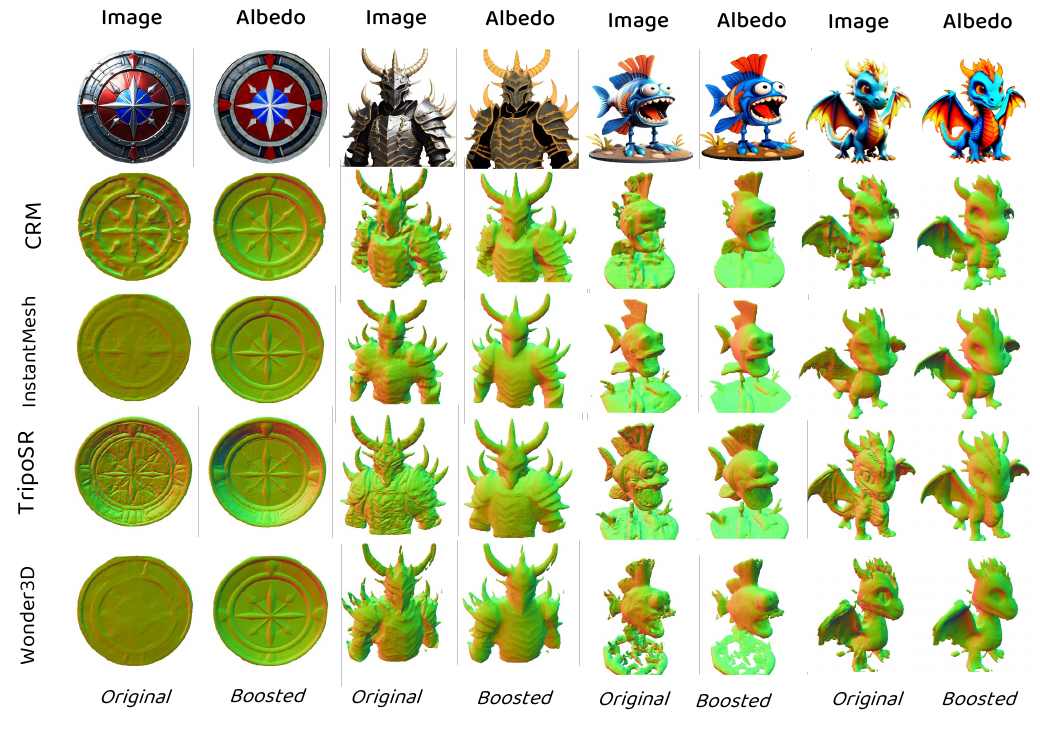}
    \vspace{-12pt} 
    \caption{\small \textbf{Normal boosting for four different methods}. Our iterative normal refinement significantly reduces the original geometry flaws and successfully captures more intricate details aligning with the corresponding images. It's noteworthy that TripoSR inevitably predicts artificial geometry details while our method can avoid this issue.}
    \label{fig:geo_boost4}
\end{figure*}

\begin{figure}[t!]
    \centering
    \includegraphics[width=0.98\linewidth]{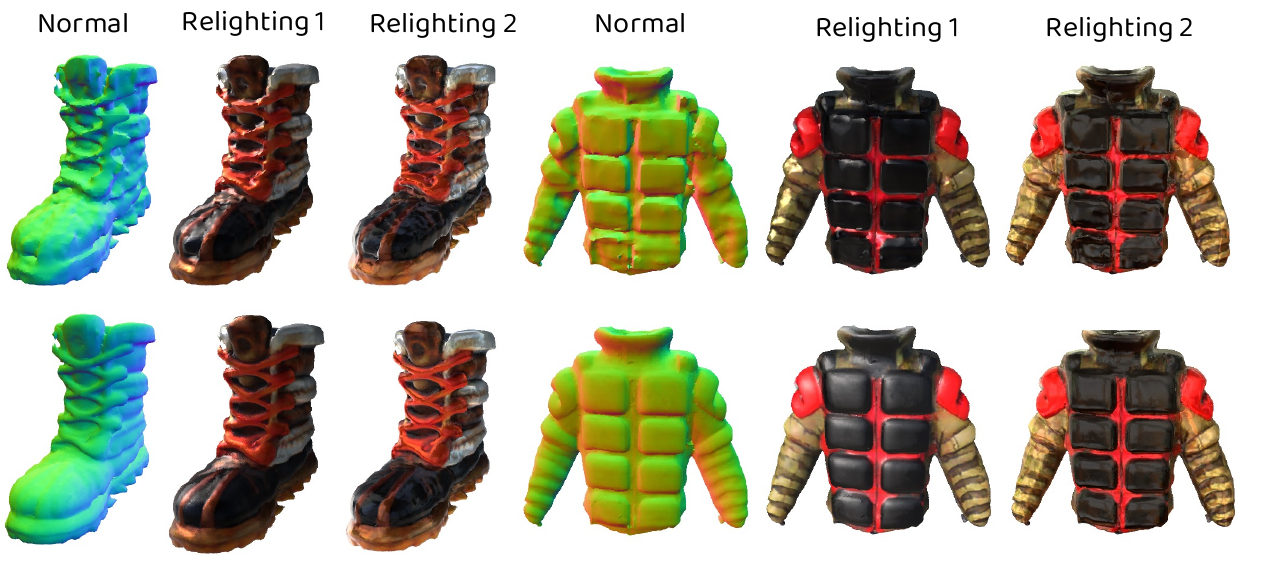} 
    \vspace{-10pt}
    \caption{\small Our refined normal maps lead to improved relighting outcomes under novel lighting environments. (\textbf{Zoom in for best view})}
    \label{fig:geo_relight}
    \vspace{-14pt}
\end{figure}

To enable the optimization of the bump map $n_b(\theta)$, we leverage our fine-tuned image-to-normal diffusion model to derive the target normal maps $n_\text{tgt}$ from the albedo maps with fine details.
Therefore, we first render multiple normal maps and the corresponding albedo maps along a series of different views.
Then, we employ the encoder $\cE$ of VAE to encode the integrated normal maps $n_f(\theta)$ and the albedo maps $a$ into latent codes $z_{n} = \cE(n_f(\theta))$ and $z_{a} = \cE(a)$, respectively.
During the inference,
The normal latent $z_{n}$ is perturbed with a random noise $\epsilon$,
and then concatenate with the albedo latent $z_a$ to predict the noise $\epsilon_\text{normal}$ following:
\begin{align}
    \epsilon_\text{normal} &= g_\text{normal} \big( z_n + \epsilon(t_0) \parallel z_a; \mathbf{s}_\emptyset, t_0 \big),
\end{align}
where $t_0$ represents the initial timestep selected to balance the information from the original normal and the priors from the image-to-normal diffusion model.
Subsequently, the target latent code $z_{n, \text{tgt}}$ can be obtained and then feed into the decoder $\cD$ of VAE for the target normal map $n_\text{tgt} = \cD(z_{n, \text{tgt}})$,
which aligns with the albedo map and thus contains enough intricate geometry details.
Regarding the target normal map $n_\text{tgt}$ as the pseudo-ground truth, we optimize the bump map $n_b(\theta)$ via a pixel-wise MSE loss:
\begin{equation}
    L_{MSE}=\Vert n_f(\theta) - n_\text{tgt} \Vert^2_2
\end{equation}

\section{Experiments}

\subsection{Implementation Details}



\begin{figure}[t!]
    \centering
    \includegraphics[width=0.95\linewidth]{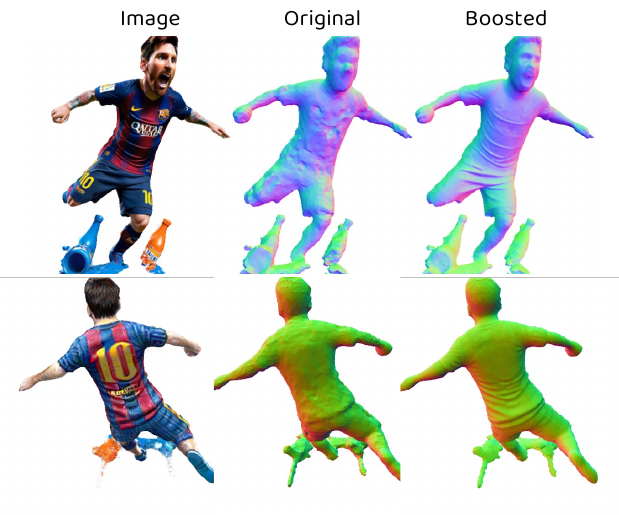}
    \vspace{-20pt}
    \caption{\small \textbf{Normal boosting on DreamCraft3D}. Our iterative normal refinement also shows its effectiveness on typical 3D objects generated by the prominent method DreamCraft3D.}
    \label{fig:geo_boost1}
    \vspace{-10pt}
\end{figure}

\subsubsection{The fine-tuning of Stable Diffusion}
\label{sec:imp_d}
The Stable Diffusion-V2.1-base model is selected as our base model for fine-tuning.
During the fine-tuning, 
we freeze the VAE and only fine-tune the U-Net using the standard denoising diffusion objective: $\mathcal{L}_\text{task} = \|\epsilon - \hat{\epsilon}_\text{task}\|_2^2$, where $\epsilon \sim \mathcal{N}(0, I)$ is a random noise map.
As mentioned in Section~\ref{sec:an_est},
the input channels of the first convolution layer inside U-Net are duplicated to empower the desirable image-to-image translation ability.
During the training, we zero-initialize the weight for the duplicated channels in the input layers and train our model on the HyperSim \cite{hypersim}, a synthetic indoor-scene dataset containing ground truth albedo and normal map.
The fine-tuning on albedo and normal maps takes 16 hours and 22 hours respectively on a single NVIDIA Tesla A100 GPU.

For the albedo estimation, we observe degenerate results on object images owing to the color space gap between indoor-level and object-level data.
To address this issue,
we further fine-tune the albedo estimation model on the Objaverse~\cite{deitke2023objaverse} dataset to align the color space of the model output to the object-level data,
which roughly requires 28 hours of training on 4 A100 GPUs.
Importantly,
we empirically find direct fine-tuning on the Objaverse dataset is insufficient to remove strong lighting effects, such as highlights and shadows, from the input RGB images.
In contrast, 
the fine-tuned image-to-normal diffusion model demonstrates superior performance on object-level data,
successfully recovering intricate normal maps from the object images.

\subsubsection{Methods for boosting}

We select four different reconstruction-based methods as the base models to boost:
\textbf{Wonder3D}~\cite{long2023wonder3d} generates 6-view images and normal maps that are fused to a textured 3D mesh via NeuS~\cite{wang2021neus};
\textbf{TripoSR}~\cite{tochilkin2024triposr} builds upon LRM structure but affords substantial improvements in model design and training processes.
While
\textbf{CRM}~\cite{wang2024crm} predicts 6 orthographic images and then employs a convolutional U-Net for 3D reconstructions,
\textbf{InstantMesh}~\cite{xu2024instantmesh} builds a purely transformer-based reconstruction architecture, offering superior flexibility and training scalability.



\subsection{Normal boosting}
In Figure~\ref{fig:geo_boost4}, we present a visual comparison of normal boosting results for four distinct base models, accompanied by the input image and our estimated albedo map at the top. As illustrated, the base methods CRM and Wonder3D yield unsatisfactory object geometries, plagued by numerous flaws, whereas InstantMesh tends to reconstruct 3D meshes that lack essential geometry details.
After our boosting, the resulting normal maps exhibit a significant reduction in geometry flaws and effectively capture more intricate details aligning with the corresponding images.
It's noteworthy that TripoSR is prone to predict more artificial geometry details but ours can successfully avoid such a dilemma.
Figure~\ref{fig:geo_relight} provides further validation of our normal boosting results through relighting experiments, wherein it is evident that the generated PBR materials yield satisfactory relighting outcomes only when combined with the boosted normal maps.
Furthermore, as shown in Figure~\ref{fig:geo_boost1},
our method is also capable of boosting the normal maps generated by DreamCraft3D~\cite{sun2023dreamcraft3d}, a prominent optimization-based approach for synthesizing 3D object meshes from single images.

\begin{figure}[t!]
    \centering
    \includegraphics[width=0.95\linewidth]{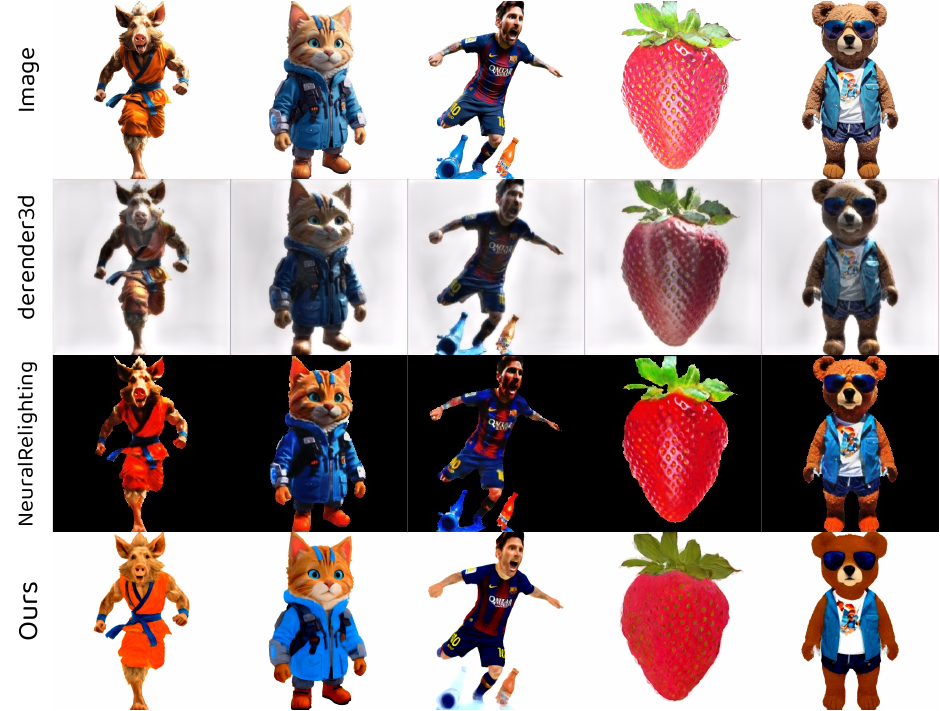}
    \vspace{-10pt}
    \caption{\small \textbf{Qualitative comparison of albedo estimation}. Regarding albedo estimation from the single image, our fine-tuned diffusion model outperforms two strong baselines on in-the-wild testing cases.}
    \label{fig:albedo_compare}
    \vspace{-10pt}
\end{figure}


\subsection{Qualitative Comparison to Baselines}
We compare our albedo estimation module with two strong baselines~\cite{sang2020single,wang2022rodin} aiming to recover the albedo \textcolor{black}{map} from the given single image.
Unlike baseline methods, our method is able to derive albedo maps that effectively eliminate strong lighting effects as shown in Figure~\ref{fig:albedo_compare}.
We also try to compare our PBR material results with baselines enabling PBR material generation.
In the absence of prior work focused on material generation for reconstruction-based image-to-3D methods, we opt to compare our material generation results with those of two representative text-to-3D approaches, Fantasia3D~\cite{chen2023fantasia3d} and RichDreamer~\cite{qiu2023richdreamer}.
Figure~\ref{fig:pbr_compare} shows that Fantasia3D fails to exclude highlights or shadows from the obtained albedo maps, whereas RichDreamer incorrectly assigns geometry details to the variations in the metalness and roughness maps, leading to unrealistic relighting results under various novel illuminations.
Thanks to our image-to-albedo diffusion model and 3D semantic masks,
we can generate high-quality PBR materials that more accurately conform to the requirements of real-world 3D content creation workflows. Moreover, our method also supports flexible material editing as shown in Figure~\ref{fig:supp_edit}.

\subsection{User Study}
We conduct an experiment involving 20 diverse 3D objects across 4 base models, totaling 80 pairwise comparisons, to evaluate the perceptual quality enhancement of our boosted results relative to prior work. For each comparison, participants will view the input image, original and boosted normal maps, and relighting outcomes pre- and post-boosting. The presentation order is randomized to maintain the questionnaire's integrity. Participants assess the quality of normal maps and the naturality of relighting, selecting the result superior in visual fidelity and realism. Feedback from 60 participants is collected, and preference ratios are computed for four different base models against our boosting model as shown in Table~\ref{tab:userstudy}.
\begin{table}[t!]
\caption{\small \textbf{User study}. The ratios show users' preference towards the perceptual quality of results generated by base models and boosting models.}
\label{tab:userstudy}
\vspace{-12.5pt}
\begin{tabular}{l ccccc}
\toprule
Method &CRM &\small{InstantMesh} &TripoSR &Wonder3D &Total \\
\midrule
base (\%) &9.55 &14.25 &17.46 &25.50 &16.49\\
boosting (\%) &90.45 &85.75 &82.54 &74.50 &83.51\\
\bottomrule
\end{tabular}
\vspace{-12pt}
\end{table}

\subsection{Usability Study}
We assess the effectiveness of our method through a usability study with two professional artists and eight non-expert Internet users unfamiliar with 3D creation. Participants can generate 3D objects using a base Image-to-3D model and enhance them with our tool. The general agreement among participants is that our tool substantially improves 3D object quality. However, one artist notes that the generated objects are incompatible with their required format, as the base models only produce triangle meshes. Additionally, seven participants express dissatisfaction with the lengthy generation process, which takes 25 minutes for base models and 5 minutes for our boosting. We anticipate future research to develop more efficient image-to-3D generative models. Regarding the boosting process, the primary time expenditure is attributed to the interactive phase with the SAM model for 3D semantic masks. We foresee the potential for streamlining this step with advanced 3D models capable of direct mask prediction.

\subsection{Ablation Study}
\subsubsection{Multi-view albedo estimation}
\begin{figure}[t!]
    \centering
    \includegraphics[width=0.95\linewidth]{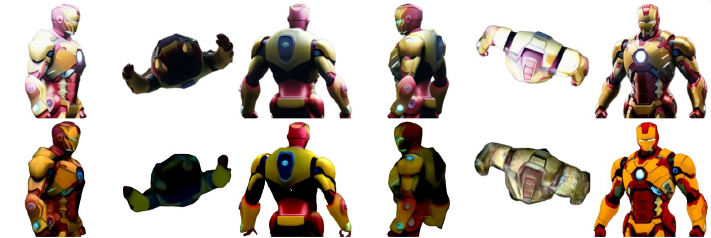}
    \vspace{-10pt}
    \caption{\small Naively applying our image-to-albedo diffusion model leads to degenerate and inconsistent albedo maps. }
    \label{fig:inconsistency}
    \vspace{-15pt}
\end{figure}

As discussed in Section~\ref{sec:pbr},
a naive application of our image-to-albedo diffusion model to multi-view RGB images would yield inconsistent albedo maps.
We show the derived albedo maps from such an inferior solution in Figure~\ref{fig:inconsistency}.

\subsubsection{Further fine-tuning on the Objaverse dataset}
\begin{figure}[t!]
    \centering
    \includegraphics[width=0.95\linewidth]{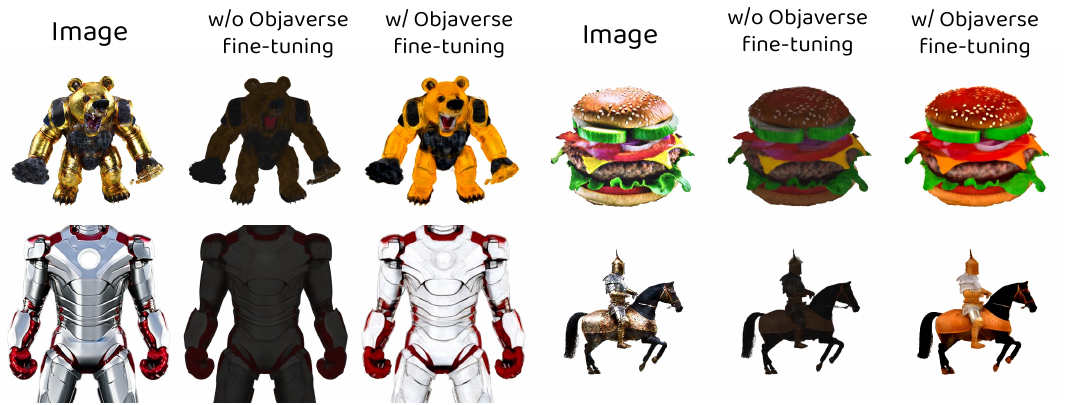}
    \vspace{-10pt}
    \caption{\small After fine-tuned on the HyperSim, our image-to-albedo diffusion model is prone to predict unnatural darker albedo \textcolor{black}{maps}, but the derived albedo turns much better after additional fine-tuning on the Objaverse.}
    \label{fig:obj}
    \vspace{-15pt}
\end{figure}

After the fine-tuning on the HyperSim dataset~\cite{hypersim}, our image-to-albedo diffusion model demonstrates suboptimal performance on object data, as illustrated in Figure~\ref{fig:obj}. The albedo comparison presented therein underscores the importance of additional fine-tuning on the Objaverse dataset~\cite{deitke2023objaverse}.

\subsection{Additional Results}
Our boosting method functions as a post-processing pipeline, independent from the 3D generation process.
To further validate the effectiveness and generalizability of our approach, we conduct additional experiments on Era3D~\cite{li2024era3d}, an advanced version of Wonder3D. The results of these tests continue to demonstrate the efficacy of our method\textcolor{black}{, as depicted in Figure 3 and Figure 4 of the supplementary material.}
\textcolor{black}{Moreover, our approach successfully enhances geometry and produces realistic PBR materials for objects crafted by professional 3D artists and featuring complex geometry and realistic appearance, which are sourced from the Objaverse-XL dataset~\cite{objaverseXL}, as illustrated in Figure~\ref{fig:add_normal}, Figure~\ref{fig:add_pbr}, and the supplementary material's Figure 1 and Figure 2.}

\section{Conclusion and Limitations}
\subsection{Conclusion}
This paper presents a novel framework for enhancing existing single image-to-3D generation methods with high-fidelity PBR materials.
Our approach involves two key components. Firstly, we adapt the Stable Diffusion model to infer albedo maps from single images and leverage powerful VLMs to derive plausible values for metalness and roughness terms. Subsequently, we augment the original texture maps with relightable PBR materials, thereby enabling realistic relighting under novel illumination conditions. Secondly, we design an iterative normal refinement module to enhance the original flawed normal maps with learnable bump maps. As a result, our refined normal maps exhibit intricate geometry details and improved alignment with the corresponding RGB images. We believe that our boosting scheme has the potential to significantly accelerate the development of single image-to-3D generation techniques.

\subsection{Limitations}
Despite the superior capability, our boosting model still exhibits certain limitations.
The image-to-albedo diffusion model introduces inherent randomness in albedo prediction, resulting in a color discrepancy between RGB images and their corresponding albedo maps. Additionally, the model's accuracy is contingent upon dataset constraints and diffusion prior performance, which does not ensure precise estimations for all images. The image-to-normal diffusion model similarly suffers from these issues. Furthermore, our pipeline's optimization of bump maps through albedo-to-normal map prediction is not wholly logical, particularly for monochromatic objects where albedo maps simplify to color blocks, devoid of geometric information, leading to the image-to-normal prediction model's failure.

\begin{acks}
This work is funded in part by the National Key R\&D Program of China (2022ZD0160201), and Shanghai Artificial Intelligence Laboratory.
\end{acks}

\bibliographystyle{ACM-Reference-Format}
\bibliography{bibliography}
\clearpage 
\appendix
\begin{figure*}[t!]
    \centering
    \includegraphics[width=0.9\linewidth]{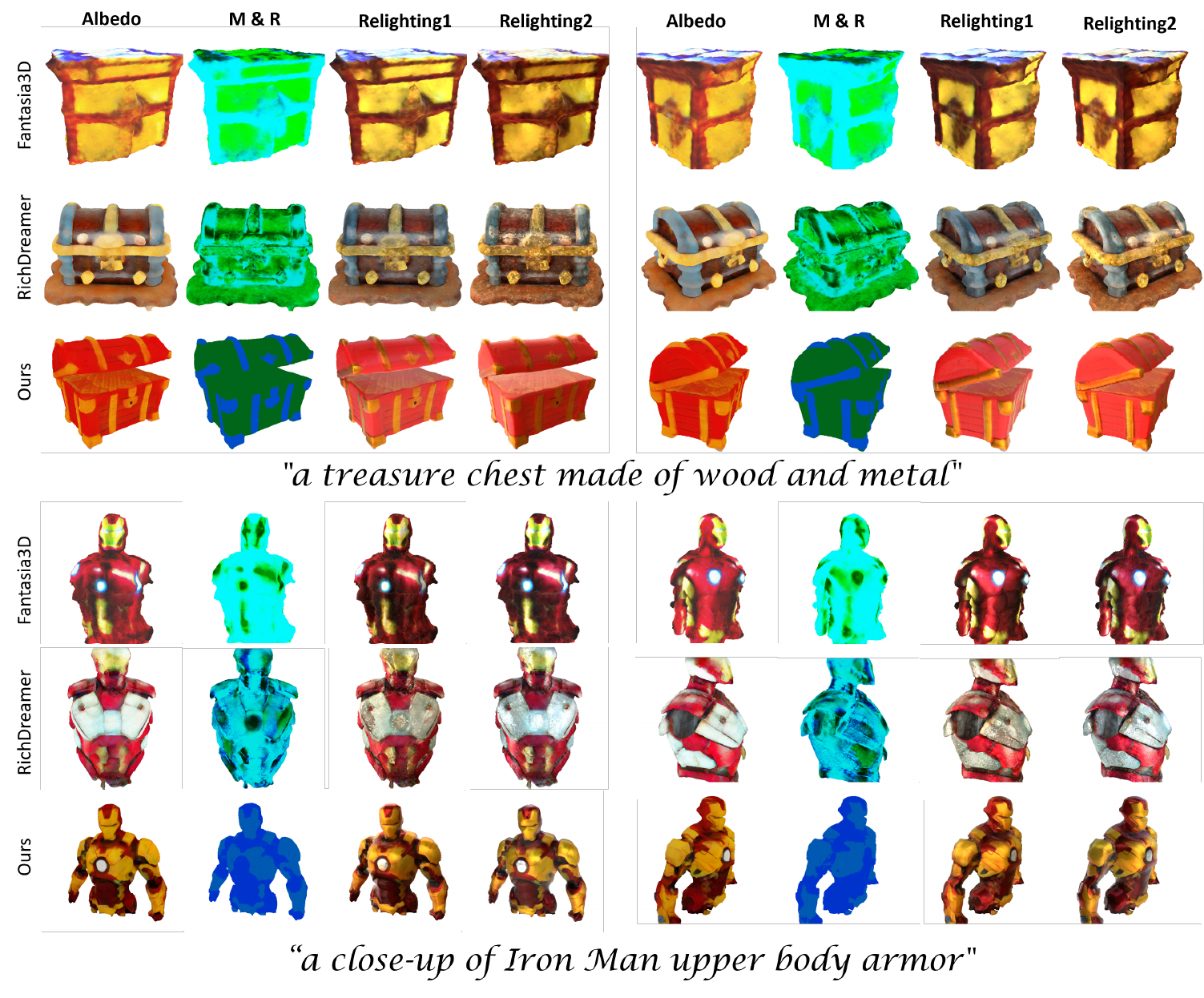}
    \vspace{-5pt} 
    \caption{\small \textbf{Qualitative comparison of material generation}. While baseline methods Fantasia3D and RichDreamer struggle to eliminate highlights or shadows from albedo \textcolor{black}{maps}, our generated PBR materials effectively circumvent this hurdle and present more natural relighting results under various illuminations. }
    \label{fig:pbr_compare}
    \vspace{0pt}
\end{figure*}

\begin{figure*}[t!]
    \centering
    \includegraphics[width=0.95\linewidth]{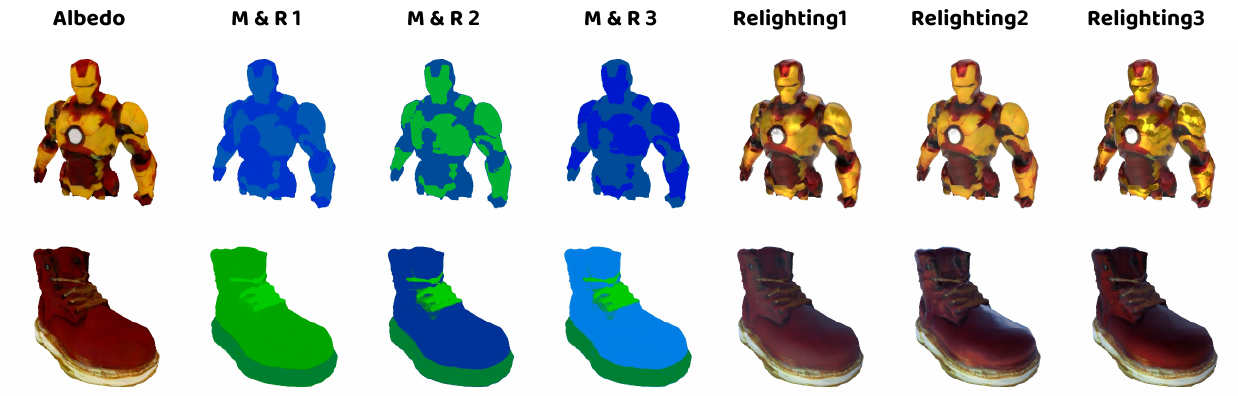}
    \caption{\small Our method empowers flexible editing on PBR materials. }
    \label{fig:supp_edit}
\end{figure*}

\begin{figure*}[t!]
    \centering
    \includegraphics[width=0.95\linewidth]{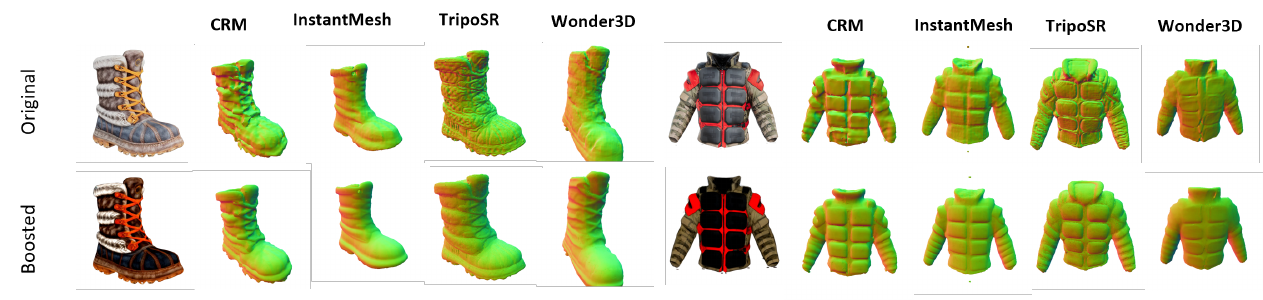}
    \vspace{-5pt}
    \caption{\small Additional results demonstrate the effectiveness of our iterative normal refinement for four different image-to-3D methods. }
    \label{fig:edit}
    \vspace{10pt}
\end{figure*}

\begin{figure*}[t!]
    \centering
    \includegraphics[width=0.95\linewidth]{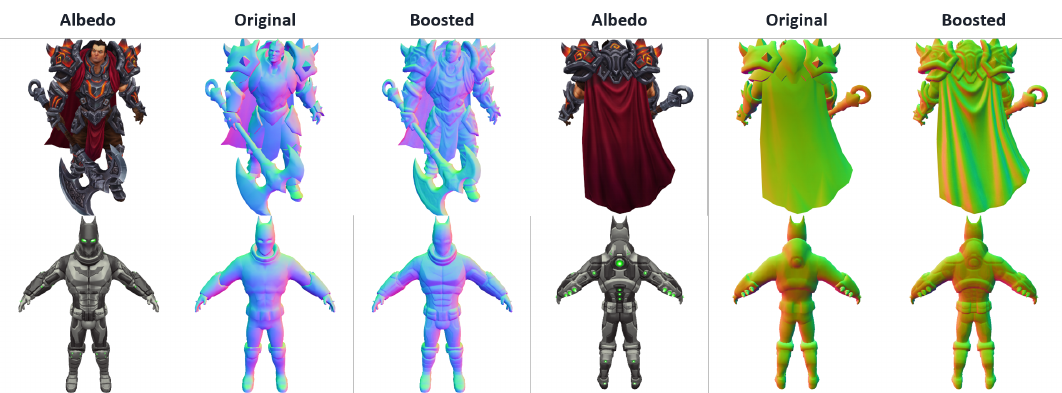} 
    \caption{\small \textbf{Normal boosting results on artist-crafted objects}. Our iterative normal refinement can also boost the normal of 3D meshes made by professional 3D artists.}
    \label{fig:add_normal}
    \vspace{10pt}
\end{figure*}

\begin{figure*}[t!]
    \centering
    \includegraphics[width=0.95\linewidth]{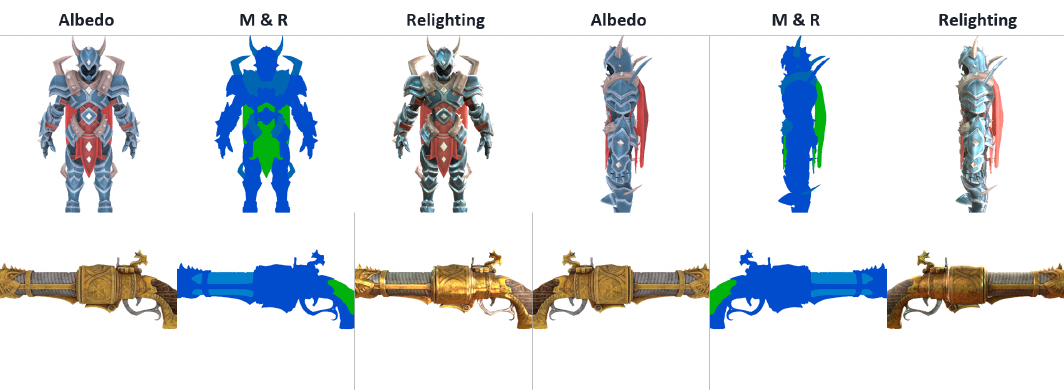} 
    \vspace{-10pt} 
    \caption{\small
    \textbf{PBR material generation results on artist-crafted objects}. It is noteworthy that these objects equipped with our generated PBR material present natural relighting results.
    }
    \label{fig:add_pbr}
    \vspace{-0pt}
\end{figure*}

\end{document}